\pdfoutput=1

\documentclass[11pt]{article}

\usepackage{acl}

\usepackage{times}
\usepackage{latexsym}
\usepackage{graphicx}
\usepackage{float}
\usepackage{CJKutf8}

\usepackage[T1]{fontenc}

\usepackage[utf8]{inputenc}

\usepackage{microtype}

\title{A Survey in Automatic Irony Processing: Linguistic, Cognitive, and Multi-X Perspectives}

\author{Qingcheng Zeng \\
  Department of Linguistics \\
  Northwestern University \\
  \texttt{qingchengzeng@outlook.com} \\\And
  An-Ran Li \\
  Department of Chinese and Bilingual Studies \\
  The Hong Kong Polytechnic University \\
  \texttt{an-ran.li@connect.polyu.hk} \\}

\begin{document}
\maketitle
\begin{abstract}
Irony is a ubiquitous figurative language in daily communication. Previously, many researchers have approached irony from linguistic, cognitive science, and computational aspects. Recently, some progress have been witnessed in automatic irony processing due to the rapid development in deep neural models in natural language processing (NLP). In this paper, we will provide a comprehensive overview of computational irony, insights from linguistic theory and cognitive science, as well as its interactions with downstream NLP tasks and newly proposed multi-X irony processing perspectives.
\end{abstract}

\section{Introduction}
Irony, which generally refers to the expressions that have opposite literal meanings to real meanings, is a representative rhetoric device in human languages \cite{li2020method}. For example, in sentences \emph{I just love when you test my patience!!}, and \emph{Had no sleep and have got school now}, compared to their literal meanings, both sentences are conveying reverse meanings and emotions, which mean \emph{not love} and \emph{not happy}. People are using ironies to communicate their affective states more implicitly or explicitly, to strengthen the claims depending on the needs, which contribute to neologism across human languages. However, the inner incongruity makes ironic expressions harder for machines to understand. Consequently, accurate irony processing systems are essential and challenging for downstream tasks and natural language understanding (NLU) research.

Previously, many researchers approached irony processing from various perspectives. The last survey in computational irony was published seven years ago \cite{wallace2015computational} and mainly focused on irony detection and pragmatic context model. Therefore, a systematic and comprehensive survey on automatic irony processing remains absent, encouraging us to focus on a review of advances in irony processing, from traditional machine learning, recurrent neural networks (RNNs) methods, to deeper pretrained language models (PLMs) throughout. Besides, \newcite{del2020review} revealed the great imbalance in figurative language research, in which sarcasm was dominant and twice as much as irony research. With this review, We aim to encourage a balanced and equal research environment in figurative languages.

Moreover, the design and further improvement in present irony processing systems should be concatenated with both the theoretical accounts in irony theories and its role in communication, followed by downstream applications in NLP tasks. Generally, most researchers approached irony processing as a single irony detection branch and three main shared tasks \cite{ghanem2019idat,van2018semeval,ortega2019overview} all focused on irony detection, whereas other aspects are constantly under-explored. And theoretical-informed or cognitive-informed discussions are rarely seen in irony research compared to tiny amendments in neural networks' architectures.

In this paper, we aim to offer a comprehensive review in irony processing from machine learning, linguistic theory, cognitive science, and newly proposed multi-X perspectives and to evaluate irony processing in NLP applications. The remaining sections are organized as follows. In the second part, we will approach irony theories in linguistics and cognitive science research, and discuss the concrete differences between irony and sarcasm. Then, we will review irony datasets in world languages and discuss potential problems in annotation schemes throughout. In the fourth part we will retrospect research progress in automatic irony processing, including traditional feature engineering, neural network architectures, to PLMs and relatively under-explored research fields in irony processing. Finally, we will discuss irony's interactions with downstream NLP tasks like sentiment analysis and opinion mining, and multi-X perspectives for further development in computational irony research.

\section{Theoretical Research in Irony}
\subsection{Irony Theories}
Various definitions have been given to irony. Early studies suggested that irony is the expression whose real meaning is contradictory to its literal meaning \cite{grice1975logic}. The Merriam-Webster Dictionary, The Oxford English Dictionary, and The Collins English Dictionary all adopted this definition and used the words "opposite" or "contrary" to explain the relationship between the literal and contextual meanings of irony.

However, more research into various types of ironic examples revealed that the contextual meaning of irony does not have to be "opposite" or "contrary" to the literal one. According to \newcite{sperber1986relevance,wilson2012explaining}, some expressions have no "literal meaning" to be challenged because no "literal meaning" is mentioned in the context, based on which they raised \textbf{relevance theory} and the ``echoic'' concept. They considered irony as ``an echoic use of language in which the speaker tacitly dissociates herself from an attributed utterance or thought'' \cite{wilson2006pragmatics}. That is, if the "echoic use" is incongruous in some ways, the expression can be ironic. Based on this theory, \newcite{neto1998onni} put forward that there are some ``echo-markers'' like \emph{definitely}, \emph{really}, and \emph{indeed}.

\newcite{li2020method} provided instances to show that "incongruity" does not have to be between the literal and contextual meanings of irony in certain circumstances. They believed that irony's true nature is a psychological activity as much as a verbal representation. The speaker or listeners must finish the "reversal" process on a psychological level for it to be completed. When compared to the concepts of "echoic" and "incongruity," “reversal” is concerned not only with the results but also with the psychological processes that the speakers/listeners go through.

\subsubsection{Types of Irony}
\newcite{booth1974rhetoric} divided irony into tragic and comic irony by literary genre, as well as stable and unstable irony by determinacy. He also categorized irony into dramatic irony, situational irony, verbal irony, and rhetorical irony by the range of context it has to refer to. Researchers in computational irony paid the most attention to situational irony and verbal irony. Situational irony is a circumstance in which the outcome differs from what was expected, or a situation that contains striking contrasts (Imagine a situation that a lifeguard is saved from drowning). Verbal irony is the expression in which the speaker's intended meaning significantly different from the literal one. \newcite{abrams2014glossary} considered that verbal irony usually involves the explicit presentation of one attitude or evaluation, but with signals that the speaker wants to express a totally different attitude or opinion. It differs from situational irony in that it is purposely manufactured by speakers. For example, \emph{Very well, keep insulting me!}


Verbal irony is the kind to which computational linguistics researchers pay the most attention. \newcite{li2021corpus} made a further classification of them. She put forward eight kinds of reversals which are rhetorical reversal, expectation reversal, evaluation reversal, reversal of sentiment, reversal of factuality, relationship reversal, reversal from opposite pair, and reversal from satiation. The paper considered that verbal ironies can be classified by the kind of reversal they generate.

\subsection{Linguistic Features}
\subsubsection{Irony Markers and Constructions}
Although most of the studies saw irony as a pragmatics phenomenon, people also considered that it can be reflected on the verbal, grammatical, or semantic level. For example, on the verbal level, people often use words like \emph{thank}, \emph{congratulate}, \emph{welcome}, \emph{happy}, and \emph{interesting} to express ironic meanings. \newcite{laszlo2017corpus} found 15 core evaluative words which often show in ironic expressions. She generated patterns from these core evaluative words to extract ironic sentences from the corpus. For example, when the word \emph{love} is in the pattern ``NP + would/ 'd/ wouldn't + love'', it is highly possible to be an ironic expression. On a grammatical level, people often use the subjunctive when they intend to be ironic. Besides that, semantic conflict is the most direct way to express ironic meaning. The incompatibility between the main words of the proposition leads to the ridiculousness of the proposition (e.g. It's very considerate of you to make such a loud noise while I was asleep). Besides, \cite{ghosh20181} also categorized irony markers according to trope, morphosyntactic, and typographic types. 

\newcite{li2021corpus} considered that ironies are often expressed by specific ``constructions'', especially in short discourses. Larger than “core evaluative words” in \newcite{laszlo2017corpus}, the ``constructions'' mentioned in \newcite{li2021corpus} are mostly in the form of idioms or phrases. The crucial feature of them is the lack of predictability. Most of them do not have to rely on too much contextual information, they themselves can provoke the process of reversal for readers or listeners (e.g. \begin{CJK*}{UTF8}{gbsn}贵人多忘事\end{CJK*} (honorable people frequently forget things)).

\subsubsection{Irony in Communication}
Researchers claim that by using ironies, people have several kinds of intentions.

\textbf{Be polite}: According to \newcite{brown1987politeness}, when unfavorable attitudes such as resistance, criticism, and complaints are stated with irony, the threat to the listener's reputation is reduced. The irony, as stated in \newcite{giora1995irony}, is an indirect negation. People prefer to utilize indirect negation to be polite to their listeners because direct negation can generate great unhappiness;

\textbf{Ease criticisms}: As reported by \newcite{dews1995muting}, irony helps to ease the expression's evaluative function. They believe that the incompatibility between literal meaning and contextual meaning can make it difficult to articulate negative feelings. However, \newcite{toplak2000uses} argued that, while irony literally avoids conflict, it is more aggressive from the perspective of the speaker's goal;

\textbf{Self-protection}: \newcite{sperber1986relevance} proposed the "echoic" idea, which stated that irony is a detached utterance that is simply an echo of another people's thought. It's a self-protection tactic, especially when the speakers are members of marginalized groups. According to \newcite{gibbs2000irony}, the irony is an ``off-record'' statement that allows speakers to deny their true intentions and avoid being challenged; 

\textbf{Be amusing}: \newcite{gibbs2000irony} reported that when young people intend to be humorous, 50\% of their communication is ironic. It can assist people in creating a dialogue platform on which speakers and listeners can agree and communicate more easily.

\subsection{Irony and Sarcasm}
Most of the studies saw sarcasm as a subset of irony \cite{tannen2005conversational,barbe1995irony,kumon1995another,leggitt2000emotional,bowes2011sarcasm}. Sarcasm is often recognized as “a nasty, mean-spirited, or just relatively negative form of verbal irony, used on occasion to enhance the negativity expressed relative to direct, non-figurative criticism” \cite{colston2017irony}.

One of the peculiarities of sarcasm is whether or not the speakers intend to offend the listeners. \newcite{kumon1995another}, for example, believe that sarcastic irony always conveys a negative attitude and is intended to harm the object being discussed. The non-sarcastic irony, on the other hand, can communicate either a good or negative attitude, and it is rarely meant to be hurtful. \newcite{barbe1995irony} concurred that the core difference was ``hurtful''. She claimed that irony is a face-saving strategy while sarcasm is a face-threatening action. Ridicule is another feature of sarcasm. According to \newcite{lee1998differential}, sarcasm is closer to ridicule than irony. Their experiment revealed that sarcasm is directed at a single person, but irony is directed toward a large group of people. \newcite{haiman1998talk} claimed that one of the most distinguishing characteristics of sarcasm is that the literal meaning of its words is always positive. However, he did not convey his thoughts on irony. Whereas \newcite{littman1991nature} viewed this topic from another angle. While there are many various forms of ironies, they believe that there is only one type of sarcasm because "sarcasm cannot exist independently of the communication setting."

Cognitive scientists approached the difference in experimental studies. Previous research in child language acquisition \cite{glenwright2010development} reported that children understood sarcastic criticism later than they could understand the non-literal meanings of irony and sarcasm, implying different pragmatic purposes of irony and sarcasm. \newcite{filik2019difference} utilized fMRI and found out sarcasm is associated with wider activation of semantic network in human brains compared to irony. 

However, most computational linguistics researchers used irony and sarcasm interchangeably, since the boundary between these two concepts is too vague for even human beings, let alone for machines. \newcite{joshi-etal-2016-challenging} and \newcite{SULIS2016132} verified this claim from both human annotators and computational perspectives. Although in this paper we will mainly focus on the literal \textbf{irony processing} and discuss the inspirations from recent research output in sarcasm processing, we aim to encourage \textbf{unify sarcasm under the framework of irony via fine-grained annotation schemes}.


\section{Irony Datasets Perspectives}
\subsection{Irony Textual Datasets}
Some main target databases for irony processing research include social media platforms like Twitter and online shopping websites like Amazon. For example, \newcite{reyes2012making} collected a 11,861-document irony dataset based on customer reviews from several websites. Some other preliminary attempts to build irony benchmark datasets is \newcite{reyes2012humor} and \newcite{reyes2013multidimensional}, in which they used self-generated hashtag \emph{\#irony} as the gold standard and constructed 40,000-tweet and 50000-tweet datasets from Twitter respectively, each including 10,000 ironic tweets and remaining non-ironic ones. The irony benchmark dataset that is now widely used is from \newcite{van2018semeval}, consisting of 4,792 tweets and half of them were ironic. This dataset was also constructed via searching hashtags including \emph{\#irony}, \emph{\#sarcasm}, and \emph{\#not}.

There were also tremendous attempts to construct benchmark datasets in other languages. For example, \newcite{tang2014chinese} firstly built the NTU Irony Corpus including 1,005 ironic messages from Plurk, a Twitter-like social media platform, by mining specific ironic patterns and manually checking extracted messages. A recent Chinese benchmark dataset for irony detection was constructed by \newcite{xiang-etal-2020-ciron}, which includes 8,766 Weibo posts, labelled from \emph{not ironic} to \emph{strongly ironic} in a five-scale system. Besides, irony datasets in Spanish, Greek, and Italian are also widely available. A comprehensive overview of the irony datasets is listed in Table~\ref{tab:1}.
\begin{table*}[]
\resizebox{\textwidth}{!}{%
\begin{tabular}{cccccc}
\hline
\textbf{Study} & \textbf{Language} & \textbf{Data Source} & \textbf{Construction Methodology} & \textbf{Size} & \textbf{Annotation Scheme} \\ \hline
\newcite{reyes2012making} & English & online websites & filtering low-star reviews & 11861 & ironic / non-ironic \\
\newcite{reyes2013multidimensional} & English & Twitter & \#irony hashtag & 40000 & ironic / non-ironic \\
\newcite{wallace-etal-2014-humans} & English & Reddit & Sub-reddit & 3020 & ironic / non-ironic \\
\newcite{van-hee-etal-2016-exploring} & English, Dutch & Twitter & \#irony, \#sarcasm, and \#not hashtags & 3000, 3179 & three-scale irony annotation \\
\newcite{van2018semeval} & English & Twitter & \#irony, \#sarcasm, and \#not hashtags & 4792 & four-scale ironic types \\
\newcite{tang2014chinese} & Chinese & Plurk & pattern mining & 1005 & annotating ironic elements \\
\newcite{xiang-etal-2020-ciron} & Chinese & Weibo & pattern mining & 8766 & five-scale ironic intensity \\
\newcite{barbieri2016overview} & Italian & Twitter & keywords, hashtags, and etc. & 9410 & ironic / non-ironic \\
\newcite{cignarella2018overview} & Italian & Twitter & keywords, hashtags, and etc. & 4849 & ironic / non-ironic / sarcastic \\
\newcite{charalampakis2015detecting} & Greek & Twitter & keywords & 61427 & ironic / non-ironic \\
\newcite{ortega2019overview} & Spanish & Twitter & comments and tweets & 9000 & ironic / non-ironic \\
\newcite{ghanem2019idat} & Arabic & Twitter & keywords and hashtags & 5030 & ironic / non-ironic \\
\newcite{DBLP:journals/pdln/CorreaCSF21} & Portuguese & Twitter and news articles & keywords, hashtags, and news articles & 34306 & ironic / non-ironic \\
\newcite{vijay2018dataset} & Hindi-English code-mixed & Twitter & keywords and hashtags & 3055 & ironic / non-ironic \\ \hline
\end{tabular}%
}
\caption{Irony benchmark datasets}
\label{tab:1}
\end{table*}

\subsection{Data Source and Construction Methodology}
Twitter, as one of the most trending social platforms, is the major source of irony benchmark datasets. Given Plurk and Weibo's similarity to Twitter, online short-text social medias are almost the only origin for present datasets, which might lead to several potential problems. For example, the limitation of 140-word introduced specific bias towards short-text classification and long texts remain a problem. Besides, the judgment of irony might be highly dependent on contextual information like previous comments or retweets and one single tweet could be meaningless itself. At last, topics on social media platforms might also be highly biased towards political or sports topics. Interestingly, previous research \cite{ghosh20181} reported that for Twitter and Reddit, different ironic markers played the most important roles, further emphasizing the needs of multiple sources for robust NLU.

As for the construction methodology, most datasets adopted "keywords and hashtags" filtering strategy, and some of them followed by human annotations. With annotation schemes put aside, \newcite{doi:10.1177/2053951720972735} explored how self-generated tags could correspond to real labels correctly via a manual semantic analysis. At the worst case, only 16\% of the tagged \emph{\#sarcasm} tweets are unambiguous sarcastic tweets according to well-trained linguists, which further emphasized the necessity of human annotation.

\subsection{Annotation Schemes}
As shown in Table~\ref{tab:1}, various datasets have been labelled differently. Most datasets were annotated as binary classes, ironic versus non-ironic. Both Chinese datasets adopted different strategies by annotating ironic elements and intensity respectively. \newcite{van2018semeval} is the only one doing fine-grained labelling, into verbal irony by polarity contrast, other verbal irony, situational irony, and non-ironic.

Another annotation scheme was proposed by \newcite{cignarella-etal-2020-marking}. They annotated irony activators at the morphosyntactic level and distinguished different types of irony activation. This annotation scheme dived into syntactic information and offered information for analyzing ironical constructions.

\subsection{Future work in datasets}
A high-quality benchmark dataset is crucial to measure NLU capability and advance future research. We are calling improvements for a future benchmark dataset from following perspectives.
\begin{itemize}
    \item [1)] Diverse data sources like literature, daily conversations, and news articles should be involved, and the distribution of text length should be relatively balanced.
    \item [2)] A uniform annotation scheme and strategy is awaiting for construction. For example, situational irony is needed; ironic intensity should be labelled as reference rather than exclusive labels in \newcite{xiang-etal-2020-ciron}; a scheme to unify sarcasm and irony could be expected.
    \item [3)] Multi-X perspectives should be incorporated into the construction of datasets. For example, a multimodal and multilingual dataset could enhance irony identification in a grounded environment.
\end{itemize}

\section{Irony Processing Systems}
In this section, we will discuss the progress of irony processing systems comprehensively, organized along the development of machine learning and deep learning.

\subsection{Irony Detection}

\subsubsection{Rule-based Detection Methods}
\newcite{tang2014chinese} extracted ironic expressions based on five patterns summarized from mandarin Chinese. \newcite{li2020method} further expanded ironic constructions to more than twenty and proposed a systematic irony identification procedure (IIP). Besides Chinese, \newcite{frenda2016computational} utilized sentiment lexicons, verb morphology, quotation marks, and etc. to design an Italian irony detection model and got competitive performance with machine learning models. However, rule-based models are too complex and hard to generalize for wider applications.

\subsubsection{Supervised non-neural network era}
Most research took irony detection as a simple classification problem. Before the popularity of deep learning, feature engineering is crucial for accurate irony detection. Generally, features could be divided into several levels.

\textbf{Lexical features} Lexical features are at the foundational level of NLP features, basically divided into bags of words (BOW) sets, word form sets, and conditional n-gram probabilities \cite{van-hee-etal-2018-usually}. Representative BOW sets mainly include n-grams and character n-grams. Word form sets focus on number and frequency, such as punctuation numbers, emoticon frequencies, character repetitions, etc. Despite their easiness to get, lexical features were proved effective in much research.

\textbf{Syntactic features} Syntax is mainly quantified via parts-of-speech and named entities. After tagging, the number and frequency of both characteristics could act as features in classification models. Besides, hand-crafted syntactic features also included clash before verb tenses \cite{reyes2013multidimensional} and dependency parsing \cite{cignarella-etal-2020-multilingual}.

\textbf{Semantic features} \newcite{van-hee-etal-2018-usually} approached semantic features based on the presence or not in semantic clusters, which were trained on a irony Twitter corpus with the Word2Vec algorithm \cite{NIPS2013_9aa42b31}.

\textbf{Linguistic-motivated features} Irony processing is deeply associated with sentiments and emotions. Therefore, researchers have offered many characteristics to capture irony patterns. For example, \newcite{reyes2013multidimensional} proposed the feature of \emph{contextual imbalance}, which was quantified via measuring the semantic similarity pairwise. Generally, most features could be categorized into \emph{ambiguity} \cite{reyes2012humor} and \emph{incongruity} \cite{joshi-etal-2015-harnessing}. Take incongruity as an example, implicit incongruity was defined as a boolean feature checking containing implicit sentiment phrases or not; explicit incongruity was defined as number of times a polarity contrast appears. Theoretical research \cite{https://doi.org/10.7275/91ey-3n44} is encouraging more semantic and pragmatic features to better capture ironies. 

Features at various levels were concatenated with classifiers, including naive bayes, decision tress and support vector machines (SVMs) \cite{van-hee-etal-2018-usually} to get final classification results.

\subsubsection{Supervised Neural Network Era}
There have been irony detection tasks in various languages after 2015, among which most participants used convolutional neural networks (CNNs) and RNNs methods to detect ironical expressions. For example, in SemEval-2018 Task 3 \cite{van2018semeval}, \newcite{wu-etal-2018-thu} classified irony tweets and their ironical types in a densely connected LSTM network, together with multitask learning (MTL) objectives in the optimization. In IroSvA \cite{ortega2019overview}, \newcite{Gonzlez2019ELiRFUPVAI} utilized transformer encoders only for detecting Spanish ironical tweets. 

Besides, \newcite{ilic-etal-2018-deep} firstly utilized contextualized word representations ELMo \cite{peters-etal-2018-deep} with Bi-LSTM to detect ironies. \newcite{ZHANG20191633} enhanced irony detection with sentiment corpora based on attention Bi-LSTM RNNs, and achieved state-of-the-art results on \newcite{reyes2013multidimensional} dataset.

\begin{table*}[htb]
\resizebox{\textwidth}{!}{%
\begin{tabular}{ccccc}
\hline
\textbf{Study} & \textbf{Input Features} & \textbf{Architecture} & \textbf{Dataset} & \textbf{\begin{tabular}[c]{@{}c@{}}Performance\\ (F1 Score)\end{tabular}} \\ \hline
\newcite{reyes2013multidimensional} & hand-crafted high-level features & decision tree & \newcite{reyes2013multidimensional}* & 70 / 76 / 73 \\
\newcite{barbieri-saggion-2014-modelling} & frequency, intensity, sentiments, etc. & decision tree & \newcite{reyes2013multidimensional}* & 73 / 75 / 75 \\
\newcite{Nozza2016UnsupervisedID} & unsupervised & topic-irony model & \newcite{reyes2013multidimensional}* & 84.77 / 82.92 / 88.34 \\
\newcite{ZHANG20191633} & word embeddings & sentiment-transferred Bi-LSTM & \newcite{reyes2013multidimensional}* & 94.69 / 95.69 / 96.55 \\
\newcite{van-hee-etal-2018-usually} & lexical, syntactic and semantic features & SVMs & \newcite{van-hee-etal-2016-exploring}** & 70.11 \\
\newcite{rohanian-etal-2018-wlv} & intensity, contrast, topics, etc. & ensemble voting classifier & \newcite{van2018semeval}*** & 65.00 / 41.53 \\
\newcite{wu-etal-2018-thu} & word embeddings, POS tags, sentiments, etc. & LSTM + MTL & \newcite{van2018semeval}*** & 70.54 / 49.47 \\
\newcite{cignarella-etal-2020-multilingual} & mBERT output, autoencoders & LSTM & \newcite{van2018semeval}*** &  70.6\\
\newcite{Potamias_2020} & RoBERTa output & RCNN & \newcite{van2018semeval}*** & 80.0 \\
\newcite{Santilli2018AKA} & word space vectors, BOW sets & SVMs & \newcite{cignarella2018overview}**** & 70.00 / 52.00 \\
\newcite{Cimino2018MultitaskLI} & word embeddings & LSTM + MTL & \newcite{cignarella2018overview}**** & 73.60 / 53.00 \\
\newcite{Gonzlez2019ELiRFUPVAI} & word embeddings & transformer encoders & \newcite{ortega2019overview} & 68.32 \\ \hline
\end{tabular}%
}
\small{* Three binary classification systems were trained respectively for this dataset.}

\small{**This result was obtained on the 3000 English tweets subset.}

\small{***This dataset had two sub-tasks, identifying ironic or not and classifying ironic types. The latter two only focused on the first sub-task.}

\small{****This dataset had two sub-tasks, irony detection and further identifying sarcasm.}
\caption{Representative Irony Detection Systems}
\label{tab:2}
\end{table*}

\subsubsection{Pretraining and Fine-tuning Paradigm}
Since BERT \cite{devlin-etal-2019-bert}, the ``pretraining and fine-tuning paradigm'' has become the mainstream in NLP research due to its extraordinary capacity in dealing with contextualized information and learning general linguistic knowledge. Recent development in irony detection also witnessed the usage of PLMs. \newcite{xiang-etal-2020-ciron} released several baseline results along with the dataset, among which BERT had highest accuracy (5\% higher than Bi-LSTM methods). \newcite{Potamias_2020} proposed a recurrent convolutional neural network (RCNN)-RoBERTa \cite{https://doi.org/10.48550/arxiv.1907.11692} strategy and improved the results on the SemEval-2018 dataset by a large degree. Besides, \newcite{cignarella-etal-2020-multilingual} explored syntax-augmented irony detection with multilingual BERT (mBERT) in multilingual settings. To sum up, some representative studies in irony detection are detailed in Table~\ref{tab:2}.

\subsection{Irony Generation}
Irony generation is mostly an underexplored research field besides \newcite{https://doi.org/10.48550/arxiv.1909.06200}, in which they defined irony generation as a style transfer problem, and utilized a Seq2Seq framework \cite{Sutskever2014SequenceTS} with reinforcement learning to generate ironical counterparts from a non-ironic sentence. Concretely, they designed the overall reward as a harmonic mean of irony reward and sentiment reward, which was trying to capture the sentiment incongruity. In terms of the evaluation, besides traditional natural language generation metrics like BLEU, they also designed task-specific evaluation metrics, which shoule be further enhanced in irony and even figurative language research.

Future work in irony generation could be advanced in new PLMs and theoretical accounts. For example, no attempts were made to generate ironical expressions after generative PLMs like BART \cite{lewis-etal-2020-bart}. Controllable irony generation and its interaction with agents are interesting topics remaining for future exploration. Besides, irony theories could be further utilized. In recent research on unsupervised sarcasm generation \cite{mishra-etal-2019-modular,chakrabarty-etal-2020-r}, context incongruity, valence reversal, and semantic incongruity were merged to enhance the generation.

\section{Discussion}
\subsection{Irony for Downstream NLP Tasks}
Irony is directly associated with downstream NLU tasks like sentiment analysis and opinion mining. For example, the sentence retrieved from \newcite{filatova-2012-irony} \emph{I would recommend this book to friends who have insomnia or those who I absolutely despise.} is classified as positive by fine-tuned sentiment analysis RoBERTa model \cite{heitmann2020}, which is apparently opposite to human evaluation. Wrong sentiment judgments will potentially lead to contrary opinion mining. We suggest that irony could be further captured through introducing incongruity embedding or specific pattern matching. \newcite{joshi-etal-2015-harnessing} designed linguistic-motivated features implicit and explicit incongruity, which are inspiring for enhancing irony understanding. Consider another example task, machine translation, in which wrong translation will potentially lead to totally opposite meanings. We encourage to model discourse features \cite{voigt-jurafsky-2012-towards}, such as ironic patterns and punctuation as embeddings for robust irony translation.

In addition, we are looking forward to the research of irony and sarcasm processing in NLP for social good (NLP4SG), especially considering the strong sentiments hidden in ironies. A recent work \cite{CHIA2021102600} explored cyberbullying detection and this was a starting point to handle online harmful ironical contents.

\subsection{Multi-X Perspectives}
Recent developments in NLP and sarcasm processing encourage us to approach discourse processing from multiple angles. In this part, we will review and suggest several multi-X perspectives for irony and figurative language processing.

\subsubsection{Multimodal Irony Processing}
Linguistic interactions are not solely consisted of texts. Besides, facial expressions and speech communications are crucial to convey emotions and feelings. For example, \newcite{skrelin2020can} reported people could classify ironies based on phonetic characteristics only. Consequently, it is conceivable that multimodal methods could help with irony detection. \newcite{schifanella2016detecting} made the first attempt in multimodal sarcasm detection, in which they extracted posts from three multimodal social media platforms based on hashtags. Then they used SVMs and neural networks to prove the validity of visual information in enhancing sarcasm detection.

\newcite{castro-etal-2019-towards} made a great improvement in multimodal sarcasm detection by introducing audio features into the dataset. The experiments also verified the importance of more modalities in sarcasm processing.

Future work in multimodal irony processing should include a comprehensive multimodal irony dataset based on MUStARD dataset \cite{castro-etal-2019-towards} with more fine-grained annotation schemes. Additionally, most methods \cite{pan-etal-2020-modeling,liu-etal-2021-smile-mean} explored sarcasm by introducing inter-modality and intra-modality attention in single-stream setting. How double-stream multimodal pretrained models (MPMs) will encode and interact in complex discourse settings remains an interesting problem to solve.

\subsubsection{Multilingual Irony Processing}
To understand irony in a multilingual context is even harder due to cultural gaps. Previously listed dataset includes a Hindi-English code-mixed irony dataset \cite{vijay2018dataset}, in which they offered an example:

\begin{itemize}
    \item \textbf{Text:} The kahawat ‘old is gold’ purani hogaee. Aaj kal ki nasal kehti hai ‘gold is old’, but the old kahawat only makes sense. \#MindF \#Irony.
    \item \textbf{Translation:} The saying ‘old is gold’ is old. Today’s generation thinks ‘gold is old’ but only the old one makes sense. \#MindF \#Irony. 
\end{itemize}

\newcite{cignarella-etal-2020-multilingual} explored how mBERT performed in multiple languages' irony detection tasks separately. Given it has been proved code-switching patterns are beneficial for NLP tasks like humor, sarcasm, and hate speech detection in RNNs settings \cite{bansal-etal-2020-code}, A future direction is to merge the irony detection datasets from multiple languages (consider \newcite{karoui-etal-2017-exploring}) or even code-mixed texts, and explore how multilingual datasets could enhance irony understanding in mBERT.

\subsubsection{Multitask Irony Processing}
MTL is to make models learn several tasks simultaneously rather than independently once at a time. Recent work in figurative language processing proposed several MTL strategies to improve the performance interactively. \newcite{chauhan-etal-2020-sentiment} proposed a MTL framework to do sentiment, sarcasm, and emotion analysis simultaneously and the framework yielded better performance with the help of MTL.

Generally, we will classify figurative language into several categories like metaphors, parodies, humors, ironies, and etc. However, noted that there are not clear differences between each other, a single task figurative language processing will only focus on one particular aspect and fail to capture the interactions. Recent work \cite{https://doi.org/10.48550/arxiv.2205.03313} also verified the combination of humor and sarcasm could improve political parody detection.

PLMs could understand figurative language better than random but apparently worse than human evaluation \cite{https://doi.org/10.48550/arxiv.2204.12632}. We suggest that future work should consider domain adaptation towards figurative language as a whole via weak supervision. MTL strategy could utilize previous research in corpus linguistics, and design an appropriate proportion in summing the loss function. Besides, a unified framework to model figurative languages could be expected.

\subsubsection{Multiagent Irony Processing}
Human-like language generation is a central topic in multiagent interactive systems. Besides robots' ironical understanding, we are also curious about how robots could generate ironical expressions. Unlike transferring non-ironic sentences to ironic, multiagent irony measures the performance during the interactions in dialogues. \newcite{Ritschel2019IronyMA} improved the robots by introducing ironic expressions, which showed better user experiences in human evaluation.

Further explorations in multiagent irony could aim at better dialogue state tracking and understand when irony should be introduced.

\subsection{New Tasks: Inspiration from Sarcasm}
Compared to sarcasm, irony is rarely seen as a term in NLP conferences. Recently we have witnessed great improvements in sarcasm processing and in this part we will discuss how new tasks in sarcasm could motivate irony research.

\textbf{Data Collection} As discussed, datasets are highly dependent on hashtags as a signal to extract ironical expressions. \newcite{shmueli-etal-2020-reactive} proposed an algorithm to detect sarcastic tweets from a thread based on exterior cue tweets. A distant supervision based method for extracting ironies from platforms is crucial, given ironies in conversational contexts are central topic in the future.

\textbf{Intended and Perceived Irony} \newcite{oprea-magdy-2019-exploring} explored how author profiling affected the perceived sarcasm (manual labelling) versus the intended sarcasm (hashtags), and verified the difference between both. Further, \newcite{oprea-magdy-2020-isarcasm} introduced iSarcasm dataset which divided intended sarcasms and perceived sarcasms. The state-of-the-art sarcasm detection models performed obviously worse than human evaluation on this dataset. Future work could focus on multimodal perceived and intended irony, especially across various cultures.

\textbf{Target Identification} Sarcasm target identification was firstly proposed in \newcite{joshi-etal-2018-sarcasm}, in which sarcasm targets were classified as one target, several targets and outside. \newcite{patro-etal-2019-deep} introduced sociolinguistic features and a deep learning framework, and improved target identification by a lot. For irony processing, most ironical expressions do not equip a specific target in itself as previously discussed. However, its ironical effects are likely in dialogue or visually grounded environment, which encourages us to enhance irony datasets in aforementioned ways.

\textbf{Irony Explanation} Irony, according to the definition, have opposite real meanings to literal meanings. However, this does not mean adding a single negation could interpret ironies well. \newcite{https://doi.org/10.48550/arxiv.2203.06419} proposed a new task, sarcasm explanation in dialogue. Irony explanation might encounter more complex problems due to relatively low proportion of targets. Still, we should include irony explanation as a branch of multimodal irony processing like \newcite{https://doi.org/10.48550/arxiv.2112.04873}.

\subsection{Explainable Irony Processing}
Explainable machine learning is of interest for most researchers to uncover the blackbox. In irony processing, we are also curious about why specific expressions are recognized as ironies. \newcite{buyukbas2021explainability} explored explainability in irony detection using Shapley Additive Explanations (SHAP) and Local Interpretable Model-Agnostic Explanations (LIME) methods. Results showed that punctuations and strong words play important roles in irony detection.

For future work, we suggest using explainable methods in multimodal settings and check how different modalities act various roles in making a class label.

\section{Conclusion}
In this paper, we reviewed the development in automatic irony processing from underexplored theoretical and cognitive science to computational perspectives, and offered a comprehensive analysis in future directions. We hope that our work and thinking will encourage further interdisciplinary research between linguistics and human language technology, motivate the research interests in irony and even, figurative languages.

\section*{Acknowledgement}
This review is based on the first author's previous research proposal. We would like to express our thanks to Professor Chu-Ren Huang and Dr. Yat Mei Lee for their suggestions.

\bibliography{anthology,custom}
\bibliographystyle{acl_natbib}

\end{document}